\documentclass[runningheads]{llncs}
\usepackage{graphicx}

\usepackage{tikz}
\usepackage{comment}
\usepackage{amsmath,amssymb} 
\usepackage{color}

\usepackage[accsupp]{axessibility}  

\begin{document}
\pagestyle{headings}
\mainmatter
\def\ECCVSubNumber{5915}  

\title{IntegratedPIFu: Integrated Pixel Aligned Implicit Function for Single-view Human Reconstruction} 

\titlerunning{IntegratedPIFu}
%
\author{Kennard Yanting Chan\inst{1,2,3}\index{Chan, Kennard Yanting} \and
Guosheng Lin\inst{3} \and
Haiyu Zhao\inst{2} \and 
Weisi Lin\inst{1,3} }
\authorrunning{K. Chan et al.}
%
\institute{S-Lab, Nanyang Technological University\\ \and
SenseTime Research\\ \and 
Nanyang Technological University\\
\email{kenn0042@e.ntu.edu.sg, zhaohaiyu@sensetime.com, \{gslin,wslin\}@ntu.edu.sg }\\ 
}
\maketitle

\begin{abstract}
We propose IntegratedPIFu, a new pixel-aligned implicit model that builds on the foundation set by PIFuHD. IntegratedPIFu shows how depth and human parsing information can be predicted and capitalized upon in a pixel-aligned implicit model. In addition, IntegratedPIFu introduces depth-oriented sampling, a novel training scheme that improve any pixel-aligned implicit model’s ability to reconstruct important human features without noisy artefacts. Lastly, IntegratedPIFu presents a new architecture that, despite using less model parameters than PIFuHD, is able to improves the structural correctness of reconstructed meshes. Our results show that IntegratedPIFu significantly outperforms existing state-of-the-arts methods on single-view human reconstruction. We provide the code in our supplementary materials. Our code is available at https://github.com/kcyt/IntegratedPIFu.
\end{abstract}

\section{Introduction}
Human digitization is an important topic with applications in areas like virtual reality, game production, and  medical imaging. While high precision reconstruction of human body is already possible using high-end, multi-view capturing systems \cite{collet2015high,guo2019relightables,lombardi2018deep}, it is largely inaccessible to general consumers. Increasingly, the research community has shown interest in developing deep learning models for human digitization from simple inputs such as a single image \cite{alldieck2019learning,natsume2019siclope,varol2018bodynet,saito2020pifuhd}.

Many methods have been proposed for single-image clothed human reconstruction, but one class of approaches that have captured significant interest is pixel-aligned implicit models \cite{saito2019pifu,he2020geo,saito2020pifuhd}. These methods model a human body as an implicit function, from which a mesh can be obtained using Marching Cubes algorithm \cite{lorensen1987marching}. One recent and notable pixel-aligned implicit model is PIFuHD \cite{saito2020pifuhd} proposed by Saito \textit{et al}. Unlike previous pixel-aligned implicit models, PIFuHD is able to capture fine-level geometric details such as clothes wrinkles. 

However, pixel-aligned implicit models, including PIFuHD, are observed to be prone to problems such as depth ambiguity and generation of meshes with broken limbs \cite{hong2021stereopifu,peng2021neural}. In particular, depth ambiguity can lead to unnaturally elongated human features in the reconstructed meshes. 

In order to overcome these issues, we propose IntegratedPIFu, which is a new and improved pixel-aligned implicit model that builds on the foundations set by PIFuHD. In order to mitigate the problem of depth ambiguity and broken limbs in reconstructed human meshes, IntegratedPIFu dedicates specialized networks for depth and human parsing predictions from a single RGB image. Predicted depth information mitigates depth ambiguity by serving as a proxy for actual depth. Human parsing predictions provide structural information critical for reconstructing human limbs without breakages or gaps.

In addition, IntegratedPIFu introduces depth-oriented sampling, a novel training scheme suitable for training any pixel-aligned implicit model. We show that this scheme is able to significantly improve the model's ability to capture important human features. Furthermore, IntegratedPIFu introduces High-resolution integrator, a new architecture that ensures coarse and fine-level components in a pixel-aligned implicit model can work together in closer tandem. We later show that our new architecture can improve the structural correctness of the reconstructed meshes.

The main contributions of our work consist of the following:
\begin{enumerate}
    \item To the best of our knowledge, we are the first to introduce a method to predict and incorporate relative depth and human parse maps into pixel-aligned implicit models for single-view clothed human reconstruction.
    \item Depth-oriented sampling, a novel training scheme that replaces the original spatial sampling scheme used to train pixel-aligned implicit models. Our scheme is able to capture small but important human features, such as fingers and ears, that are often ignored by the original scheme. In addition, unlike the original scheme, depth-oriented sampling does not cause highly noticeable wavy artefacts on reconstructed meshes.
    \item High-resolution integrator, a newly coined neural architecture that enables much stronger interactions between the coarse and fine-level components in a pixel-aligned implicit model. It is also an efficient architecture that improves the structure of the reconstructed meshes despite using 39\% less model parameters than its direct predecessor. 
    
\end{enumerate}

All three of our contributions can be applied to any existing pixel-aligned implicit model.

\begin{figure}[t]
\centering
\includegraphics[width=0.9\textwidth]{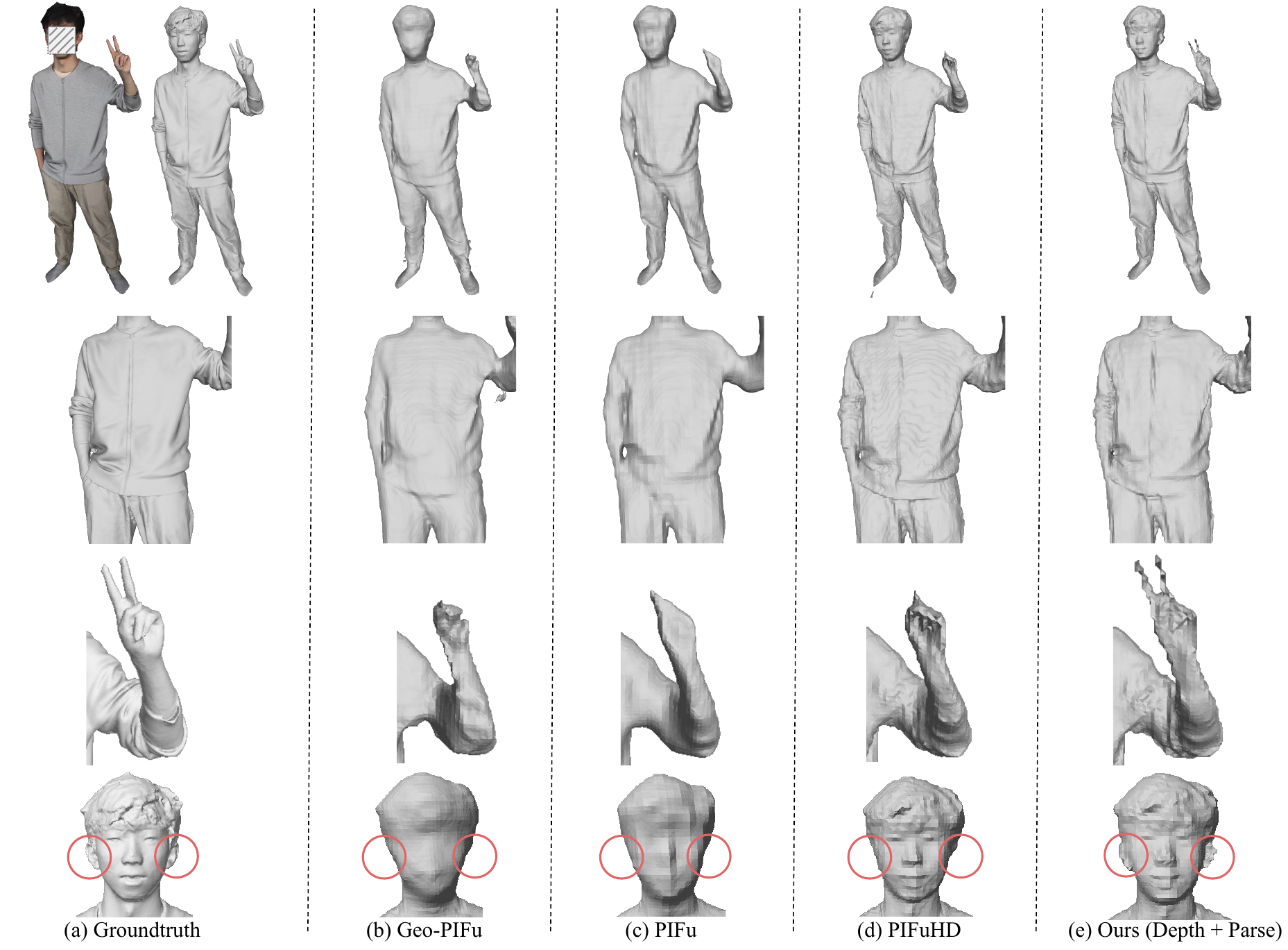}
\vspace{-4mm}
\caption{Compared with state-of-the-art methods, including (b) Geo-PIFu \cite{he2020geo}, (c) PIFu \cite{saito2019pifu}, and (d) PIFuHD \cite{saito2020pifuhd}, our proposed model can precisely reconstruct various parts of a clothed human body without noisy artefacts. Subject's face censored as required by dataset's owner}
\label{fig:firstPageSota}
\vspace{-6mm}
\end{figure}

\section{Related Work}

\subsection{3D Computer Vision Tasks}
3D Computer Vision is a overarching field that includes various diverse fields such as Point Cloud Segmentation (e.g. \cite{tang2022mpt}) and Neural Radiance Field (NeRF) (e.g. \cite{tang2023able}). In this work, we focus on Single-view Human Reconstruction, which is to generate a 3D clothed human mesh from a given input image.   

\subsection{Single-view Human Reconstruction}
Several methods have been proposed to reconstruct human body shape from a single image. The methods can be classified into parametric and non-parametric approaches. Parametric approaches, such as  \cite{kanazawa2018end,kolotouros2019convolutional,kolotouros2019learning,pang2022benchmarking,pang2024towards}, recover human body shapes by estimating the parameters of a human parametric model (e.g. SMPL \cite{loper2015smpl}) from an input image. However, these parametric methods are only able to reconstruct ``naked" (without geometry of clothes) human meshes. Bhatnagar \textit{et al.} \cite{bhatnagar2019multi} and Jiang \textit{et al.} \cite{jiang2020bcnet} additionally estimate the clothes on top of a human parametric model. However, these methods are unable to reconstruct high-fidelity geometry shapes.

Non-parametric approaches, on the other hand, do not require a human parametric model. For example, \cite{varol2018bodynet} use a 3D CNN to directly estimate a volumetric body shape from a single image. Notably, a subclass of non-parametric approaches that have garner significant interest is pixel-aligned implicit models \cite{saito2019pifu,hong2021stereopifu,he2020geo,saito2020pifuhd,chan2022s,chan2024fine,chan2024r}. One of the first pixel-aligned implicit models is the Pixel-Aligned Implicit Function (PIFu) proposed by Saito \textit{et al.} \cite{saito2019pifu}. PIFu is memory efficient and capable of producing high-fidelity 3D reconstructions of clothed human meshes. 

Building from PIFu, there have been other pixel-aligned implicit models being proposed. Examples include GeoPIFu \cite{he2020geo} and StereoPIFu \cite{hong2021stereopifu} that make use of latent voxel features and an additional view respectively to improve the quality of their reconstructed meshes. However, GeoPIFu is unable to reconstruct fine geometry details of a human body, and StereoPIFu requires an additional view and careful calibration of a binocular camera. On the other hand, PIFuHD \cite{saito2020pifuhd}, which is a multi-scale variant of PIFu and a follow-up work by Saito \textit{et al.}, is able to reconstruct very fine geometry details such as clothes wrinkles.

\subsection{PIFu and PIFuHD}
PIFu \cite{saito2019pifu} can be interpreted as having two separate parts - 1. Encoder and 2. Decoder. An illustration of the PIFU architecture is given in Fig. \ref{fig:methodDiagram} (under ``Low-Resolution PIFu"). The encoder is a 2D convolutional network that extract feature maps from a RGB image. Points are sampled from the 3D camera space (of the RGB image), and their (x,y) coordinates are used to index the feature maps to retrieve a set of feature vectors. The z value of a point, together with the feature vector retrieved by the point, would then be used by the decoder (usually a multi-layer perceptron or MLP) to predict the binary occupancy value of that point. By evenly sampling the 3D camera space and employing the Marching Cube algorithm \cite{lorensen1987marching}, a PIFu can reconstruct a 3D mesh.

PIFuHD \cite{saito2020pifuhd} improves upon PIFu by using two PIFu modules that works at different resolutions. Additionally, PIFuHD also predicts frontal and rear normal maps from a single RGB image. In PIFuHD, the first PIFu (i.e. low-res PIFu) is given a downsampled RGB image and a rear and frontal normal maps as input. The second PIFu (i.e. high-res PIFu) is given the same inputs but at the original resolution. For each point sampled from the 3D camera space, the high-res PIFu's decoder would obtain the point's corresponding feature vector from the encoder before combining that feature vector with additional features obtained from the low-res PIFu's decoder. The high-res PIFu's decoder would then predict the binary occupancy value of that point. A 3D mesh can then be generated using the same approach as PIFu. 

\subsection{Signed Distance Field}
Our depth-oriented sampling technique drew inspiration from DeepSDF \cite{park2019deepsdf}, which also employs a continuous signed distance function. But in our supplementary materials, we explain why DeepSDF, unlike DOS, is not a suitable solution for a pixel-aligned implicit model.

\section{Method}

\begin{figure}[t]
\centering
\includegraphics[width=0.9\textwidth]{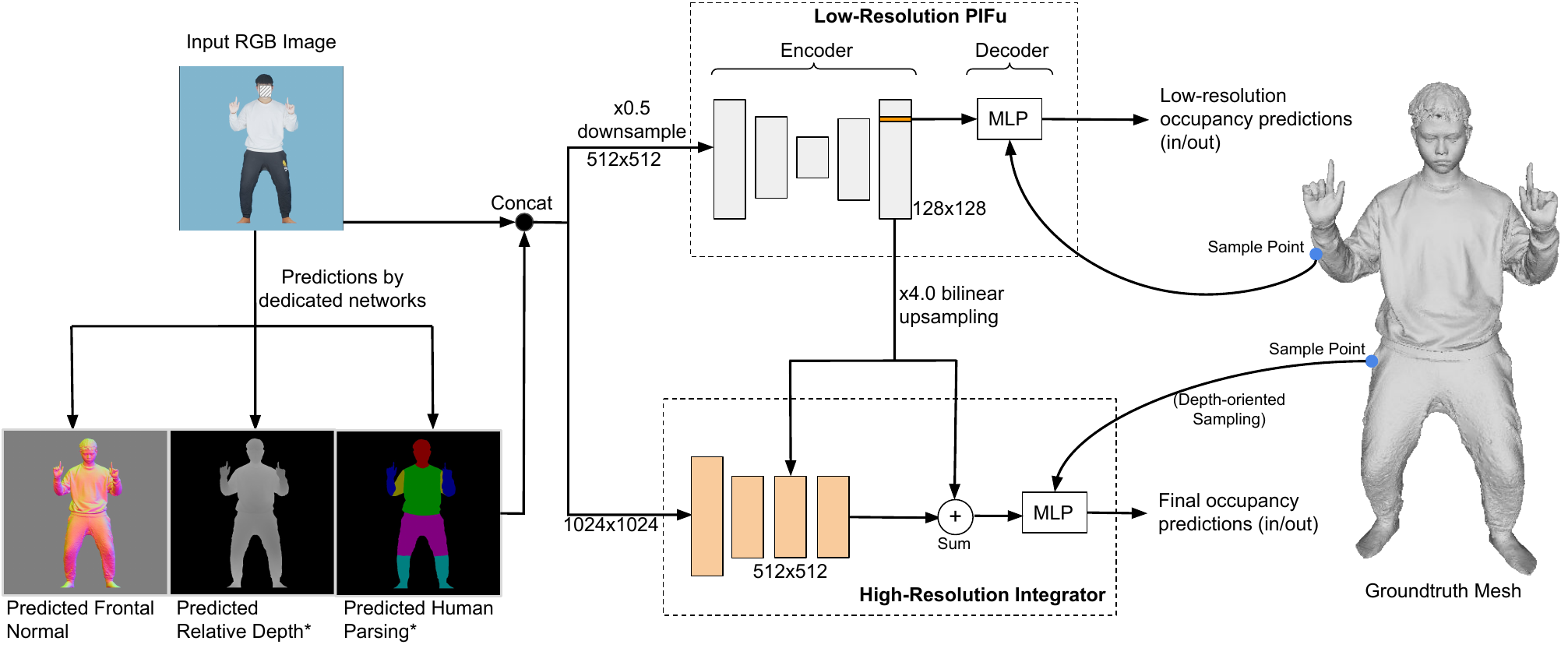}
\vspace{-4mm}
\caption{Overview of our IntegratedPIFu framework. The Low-Resolution PIFu produces coarse feature maps that are further refined by High-Resolution Integrator. The High-Resolution Integrator will then use a MLP to generate a reconstructed mesh. *We recommend the use of predicted depth and human parsing maps in the Low-Resolution PIFu, but the use of them in the High-Resolution Integrator is optional and depends heavily on the quality of the two maps.}
\label{fig:methodDiagram}
\vspace{-2mm}
\end{figure}

Our method builds on PIFuHD, which we consider to be the current state-of-the-art in this field. An overview of IntegratedPIFu is shown in Fig. \ref{fig:methodDiagram}. We train specialised networks to predict frontal normal, relative depth, and human parsing maps from a single 1024x1024 RGB image. Along with the image, these maps are downsampled (to 512x512 resolution), concatenated, and given to a PIFu. The feature maps produced by the PIFu's encoder is then upsampled bilinearly before being fed into our High-Resolution Integrator. Our High-Resolution Integrator combines the upsampled feature maps with \textbf{only} the 1024x1024 RGB image and predicted frontal normal map before producing occupancy predictions using a MLP. We train IntegratedPIFu using our newly proposed depth-oriented sampling scheme. 

We like to highlight that, unlike PIFuHD, IntegratedPIFu does not use a rear normal map as input as it tends to cause hallucinations of creases or wrinkles on the rear-side of a reconstructed mesh (More details in supplementary materials).   

In short, IntegratedPIFu offers three new contributions: 1. Incorporating depth and parsing information 2. Depth-oriented sampling 3. High-resolution integrator. We will explain the motivations for and inner workings of each of them in the following sections.

\subsection{Prediction and Incorporation of Relative Depth and Human Parse Information}

Similar to PIFuHD \cite{saito2020pifuhd}, we use a pix2pixHD \cite{wang2018high} network to predict a frontal normal map from a single RGB image. Additionally, we find that using this normal map as an additional input for relative depth and human parse prediction (carried out by specialised networks) helps to predict more accurate depth and human parse maps (shown in our supplementary materials).

\subsubsection{Depth Prediction}
In our dataset, each input RGB image is rendered from a realistic human mesh. In order to reconstruct the 3D mesh shown in a single RGB image, a pixel-aligned implicit model is implicitly required to predict the depth of each mesh vertex in the 3D camera space. This is not an easy task due to depth ambiguity in a 2D RGB image. Also, 3D mesh vertices, when being projected onto an image, can have very similar RGB values, and yet have very different depth values in the 3D space. 

Thus, we decided to partially shift the burden of depth prediction from a pixel-aligned implicit model to a separate, specialised network. This network (called depth predictor) would predict the relative depth values of mesh vertices that are visible in a RGB image (i.e. camera-facing vertices). The usefulness of incorporating relative depth information into a pixel-aligned implicit model has been soundly demonstrated in StereoPIFu \cite{hong2021stereopifu}, though in their case they used two different, carefully positioned views of the mesh to generate a highly accurate relative depth map via triangulation.

The inputs to the depth predictor is a 2D RGB image, a predicted frontal normal map, and a center indicator map. The output is a 2D relative depth map (same size as RGB image), where each pixel contains a depth value that indicates how near it is to the camera compared to a reference mesh vertex on the mesh. Our reference vertex is the front-most vertex that is located at the centre of the image, and it should always have a relative depth value of zero. We will elaborate on the center indicator map. 

The depth predictor is fundamentally a fully convolutional network with spatial invariance property. However, we require the center pixel in the relative depth map produced by depth predictor to have a value of zero (Since it is where the reference vertex is). To solve this, we introduce what we call a center indicator map, which is simply a map of zero values except its center pixel (has value of one). It has the same size as the RGB image. This helps the depth predictor to locate and predict a zero for the center pixel of all relative depth maps it generates.

The depth predictor consists of a two-staged U-Net \cite{ronneberger2015u}. Using the aforementioned inputs, the first stage U-Net would produce a coarse relative depth map. The second stage U-Net then refines the coarse relative depth map. The second U-Net is given the same inputs except that a coarse relative depth map replaces the center indicator map. In our supplementary material, we show the importance of having this two-staged process.

\subsubsection{Human Parse Prediction}
As observed by \cite{hong2021stereopifu,peng2021neural}, pixel-aligned implicit models, such as PIFu and PIFuHD, has a tendency to produce broken limbs in some of its reconstructed 3D meshes. We argue that by feeding human parsing information into a pixel-aligned implicit model, the model can better understand the general shape and location of human's limbs and parts. For example, the model should learn from a human parsing map that a normal human arm is a continuous (no gap or breakage in between) and elongated structure. 

Rather than manually labelling each pixel in every RGB image with human parsing information, we use a Self-Correction for Human Parsing model \cite{li2020self} that has been pre-trained on the Pascal-Person-Part dataset \cite{chen2014detect} to automatically generate human parsing maps from images. We find that the human parsing results, while not perfect, is reasonably correct in majority of the cases. The pre-trained model is used only to generate human parsing maps of training instances.

We treat these generated human parsing maps as the groundtruths, and use these groundtruths to train a separate network (called human parsing predictor or HPP) that would be used in IntegratedPIFu. We do this to ensure that only the necessary information from the Pascal-Person-Part dataset is transmitted to HPP. HPP is a U-Net that takes in a RGB image and predicted frontal normal map and outputs a corresponding human parsing map.

\subsection{Depth-Oriented Sampling}
Pixel-aligned implicit models need to be trained with 3D sample points. Each sample point is given a scalar label. Sample points outside the mesh are given a label of zero, and sample points inside the mesh are given a label of one. During training, a pixel-aligned implicit model will learn to assign a prediction of one to points it believes is inside the mesh and a prediction of zero to points outside of the mesh. As discussed by Saito \textit{et al.} in \cite{saito2019pifu}, the scheme that is used to generate the sample points is very important as it can adversely affect the quality of the reconstructed mesh.

\begin{figure}[t]
\centering
\includegraphics[width=0.9\textwidth]{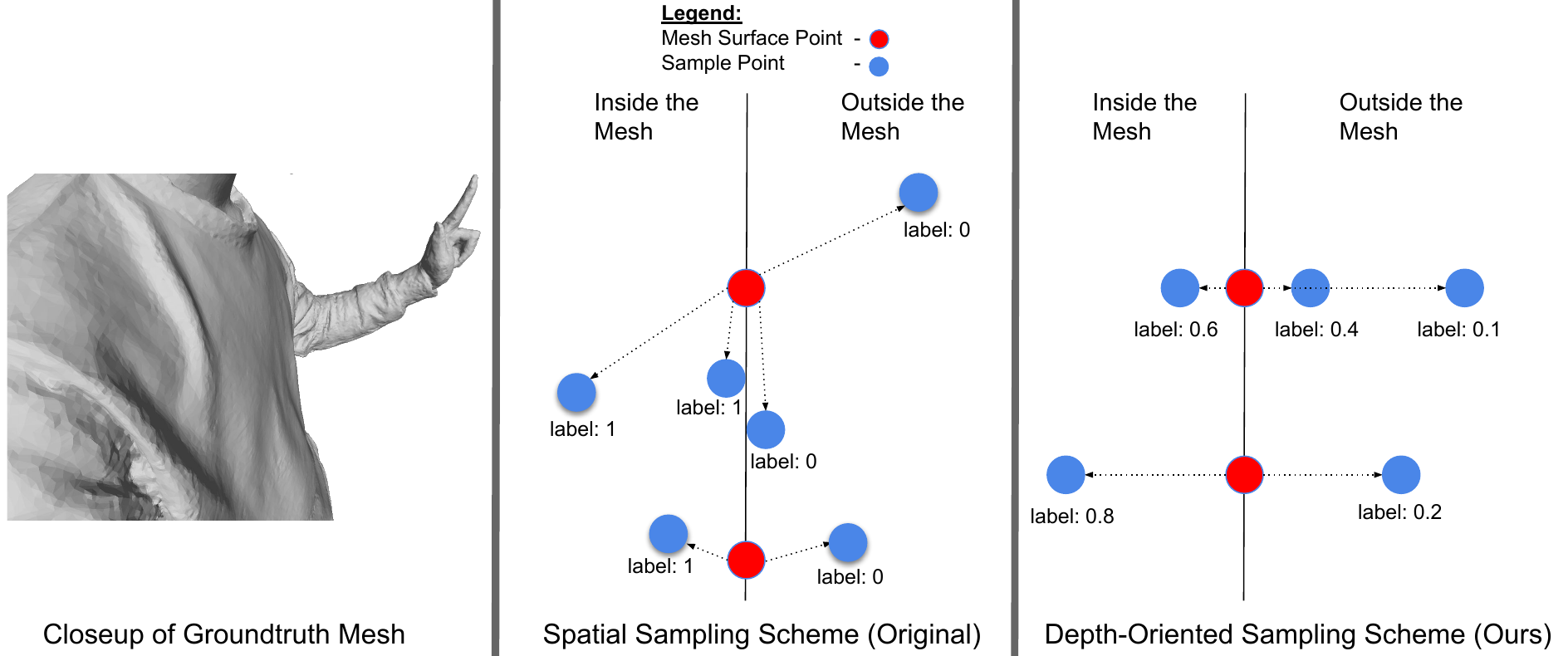}
\vspace{-4mm}
\caption{Illustration of Depth-Oriented Sampling. The red dots represent points that are exactly on the mesh surface. These red dots are then randomly displaced. The blue dots represent the possible locations of the red dots after displacement}
\label{fig:depthOrientedSamplingDiagram}
\vspace{-4mm}
\end{figure}

To the best of our knowledge, all existing pixel-aligned implicit models, including PIFu \cite{saito2019pifu}, PIFuHD \cite{saito2020pifuhd}, GeoPIFu \cite{he2020geo}, and StereoPIFu \cite{hong2021stereopifu}, use what is called the spatial sampling scheme during training. Spatial sampling scheme is the original sampling scheme developed by Saito \textit{et al.} in \cite{saito2019pifu} to generate sample points to train a PIFu. Two types of sample points exist in spatial sampling. The first type of sample points are points that are sampled directly on the mesh's surface. These points are then perturbed with normal-distributed noise (See Fig. \ref{fig:depthOrientedSamplingDiagram}). The second type of sample points are points that are uniformly sampled from a 3D camera space. Typically, the ratio of the first type of sample points to the second type is set to 16:1. Regardless of whether a sample point is of the first or second type, if it is inside a mesh, it would be given a label of one. Otherwise, it will have a label of zero. 

The first problem with spatial sampling is that if a sample point is only just slightly inside a mesh, it is given a label of one. If a similarly located sample point is just slightly outside of the mesh, then it is given a very different label of zero. On the other hand, both a sample point that is slightly outside of the mesh and a sample point that is very far away from the mesh is given the same label of zero. This problem is illustrated in Fig. \ref{fig:depthOrientedSamplingDiagram}

We argue that this scheme confuses a pixel-aligned implicit model and make it harder to train it, leading to noises in reconstructed meshes. We believe that this is the cause of noisy, wavy-like artefacts on the reconstructed mesh surfaces as seen in Fig. \ref{fig:depthOrientedComparison}b.

The second problem with spatial sampling is that it tends to ignore small but important human body features such as human ears or fingers and reconstruct meshes that have neither ears nor fingers. Small human body features like an ear have very small volume and, consequently, very few sample points would be located inside that small body feature. Instead, most sample points would be located outside of it and have a label of zero. This means that a pixel-aligned implicit model trained with spatial sampling tends to predict labels of zero for the region containing an ear (i.e. the model predicts the region to be a pure empty space rather than containing a surface of a mesh).

These two problems inspired us to develop a new sampling scheme called depth-oriented sampling (DOS). In DOS, we sample points on the mesh surface and displace them only in the z-direction (i.e. camera direction. See Fig. \ref{fig:depthOrientedSamplingDiagram}). We use a normal distribution to determine the \textbf{magnitude} of displacement. In addition, rather than limiting labels to either value of zero or one, DOS allows labels to range from 0.0 to 1.0 (i.e. soft labels). The value of the label is indicative of the distance between a particular sample point and the mesh's surface in the z-direction. A label of 0.5 indicates that the sample point is exactly on the mesh surface. A label of 0.0 indicates that the sample point is very far away from the mesh surface and is outside of the mesh body. A label of 1.0 indicates that the sample point is very far away from the mesh surface but is inside the mesh body.

This new scheme naturally solves the first problem (similarly located points being given very different labels), and it also resolves the second problem since points that are outside but in front of a small human feature (e.g. a ear) will have labels that indicates how near the surface of the human feature is, rather than completely dismissing the existence of the human feature.

\subsection{High-resolution Integrator}
PIFuHD \cite{saito2020pifuhd} consists of two PIFu modules working at different resolution levels. The low-resolution PIFu (low-PIFu) would reconstruct a coarse, general shape of the human mesh, and the high-resolution PIFu (high-PIFu) would add details onto the coarse human mesh. As aforementioned, each PIFu module consists of an image encoder (uses stacked hourglass network \cite{newell2016stacked}) and a MLP (See ``Low-Resolution PIFu" in Fig. \ref{fig:methodDiagram}).

The first problem with this is that stacked hourglass network is heavy in parameters and would take up a significant amount of GPU memory during training. Consequently, given a 1024x1024 RGB image, the high-PIFu can only be trained using image crops (as discussed in \cite{saito2020pifuhd}). Due to this, the high-PIFu does not have access to global information from the entire 1024x1024 image. Even if the high-PIFu is given enough number of image crops such that the original 1024x1024 image could be pieced together, the high-PIFu does not have information of where, in the original image, the crops are taken from. 

The first problem would not be a problem if the high-PIFu has direct and early access to information generated by the low-PIFu, and this leads us to the second problem. The low-PIFu is trained on 512x512 low-resolution images in its entirety. Thus, the low-PIFu is aware of the global information in an image. However, the high-PIFu has no opportunity to use any information from the low-PIFu until late in its pipeline. Specifically, the high-PIFu's encoder does not get to interact with information from the low-PIFu. That information is given only to the high-PIFu's MLP component. This means that the high-PIFu's image encoder does not actually know what information it should be looking for in the 1024x1024 image. Ideally, the high-PIFu's image encoder should collect complementary information that is not already captured by the low-PIFu. 

In order to solve the two problems mentioned above, we introduce a new architecture, High-resolution Integrator (HRI), to replace the high-PIFu. Similar to high-PIFu, the HRI has a encoder and a MLP. But rather than using a parameter-heavy stacked hourglass network for its encoder, the HRI's encoder adopts a shallow architecture consisting of only four convolutional layers. Thus, it consumes less GPU memory and can be trained using entire images (rather than image crops). 

A distinguishing feature of HRI is how these four layers are being used (refer to ``High-Resolution Integrator" in Fig. \ref{fig:methodDiagram}). The first and second convolutional layers are used to process the 1024x1024 high-resolution input image into a 512x512 feature map $F$. Then, using bilinear upsampling, we upscale the 128x128 feature map from the low-PIFu's encoder by a factor of 4. This upscaled feature map (denoted as $U$) is then concatenated with $F$. The concatenated feature map is then processed by the third and fourth convolutional layers. The output of the fourth convolutional layer is then added with $U$. In other words, we are using the principle of residual networks \cite{he2016deep} to train the HRI's encoder to collect only complementary information not captured by low-PIFu. Crucially, HRI's encoder directly interacts with the output from low-PIFu before producing its own output, allowing it to infer what information is missing from low-PIFu.

\section{Experiments}

\subsection{Datasets}
We train our models and other competing models on the THuman2.0 dataset \cite{tao2021function4d}, which consists of high-quality scans of Chinese human models. We remove scans or meshes that contain obvious scanning errors. Scanning errors tend to occur on the female meshes due to their free-flowing hairs. We also remove meshes that suffer from extreme self-occlusion because we do not hope to reduce the task into a guessing game. This data cleaning process is done before any of the models is trained to ensure fair comparison between the models. In all, we use 362 human meshes from the THuman2.0 dataset. We adopted a 80-20 train-test split of these meshes. For each training mesh, we render 10 RGB images at different yaw angles using a weak-perspective camera.

In addition, we also use the BUFF dataset \cite{Zhang_2017_CVPR} to evaluate the different models. No model is trained on the BUFF dataset. Systematic sampling based on sequence number is done on the BUFF dataset to obtain 93 human meshes for testing the different models. Systematic sampling ensures that, for the same human subject, meshes of different poses are obtained and repeated poses are avoided. More implementation details is in our supplementary materials.

\vspace{-2mm}

\subsection{Comparison with State-of-the-art}

We compare our method against the existing methods on single-view clothed human reconstruction qualitatively and then quantitatively. As in \cite{saito2019pifu,saito2020pifuhd,he2020geo}, the metrics in our quantitative evaluation include Chamfer distance (CD), Point-to-Surface (P2S), and Normal reprojection error (Normal). We observe that CD and P2S tends to contain a degree of noise due to the lack of depth information in a 2D RGB image. Specifically, although depth ambiguity in RGB image allows multiple plausible 3D reconstruction outputs, CD and P2S assume that only one 3D mesh output with very specific depth values for each mesh vertex can be correct.  

Existing methods that we compared against include PIFu \cite{saito2019pifu}, Geo-PIFu \cite{he2020geo}, and PIFuHD \cite{saito2020pifuhd}. We did not include parametric approaches, such as DeepHuman \cite{zheng2019deephuman} and Tex2shape \cite{alldieck2019tex2shape}, because \cite{he2020geo}, \cite{saito2020pifuhd}, and \cite{hong2021stereopifu} have all shown that pixel-aligned implicit approaches outperformed parametric ones. 

In addition, we did not compare our method with hybrid approaches (e.g. ARCH \cite{huang2020arch} and ARCH++ \cite{he2021arch++}) that combine parametric approaches with pixel-aligned implicit function approaches because of two reasons. The first reason is that ARCH and ARCH++ are solving the problem of generating animatable meshes, which is a different problem from ours. The second reason is that these approaches make use of information from an additional dataset. ARCH and ARCH++ both make use of parametric human body models estimated by DenseRaC \cite{xu2019denserac} using 4,400 human scans from the CAESAR dataset \cite{robinette2002civilian}. Furthermore, our approach can be easily slotted into ARCH and ARCH++ as both methods make use of a pixel-aligned implicit model, and our approach is centred around techniques to improve a pixel-aligned implicit model.

\vspace{-3mm}

\subsubsection{Qualitative Evaluation}
\begin{figure}[t]
\centering
\includegraphics[width=0.9\textwidth]{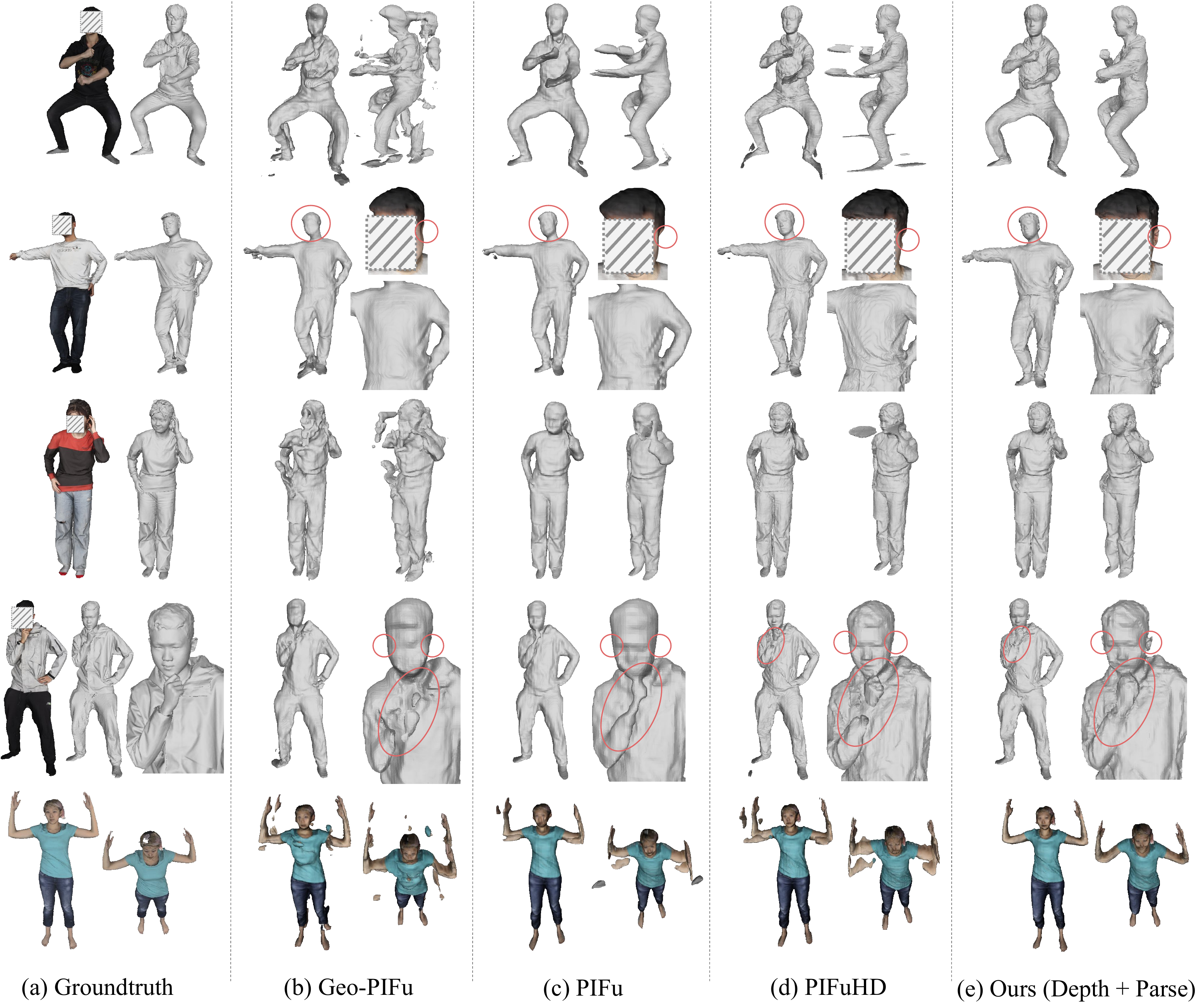}
\vspace{-4mm}
\caption{Qualitative evaluation with SOTA methods, including (b) Geo-PIFu \cite{he2020geo}, (c) PIFu \cite{saito2019pifu}, and (d) PIFuHD \cite{saito2020pifuhd}. The input RGB image is shown as the first object in each row. For each method, we show the frontal view and an alternative view. For the last row, the reconstructed meshes are colored by projecting the RGB values from the input image onto the generated meshes. The coloring serves as visual aids only.}
\label{fig:secondPageSota}
\vspace{-6mm}
\end{figure}

We present a qualitative comparison of our IntegratedPIFu (i.e. HRI + DOS + predicted depth and parsing maps) with the other existing methods in Fig.\ref{fig:firstPageSota} and Fig.\ref{fig:secondPageSota}. In Fig.\ref{fig:firstPageSota}, we observe that only PIFuHD and our approach are able to reconstruct human facial features with great details. However,
PIFuHD tends to suffer from wavy artefacts. Moreover, only our approach can reconstruct small but important human body parts like fingers and ears.  

In Fig.\ref{fig:secondPageSota}, we observe that Geo-PIFu, PIFu, and PIFuHD tend to generate floating artefacts that are very clearly visible from the side-view or top view. In contrast, IntegratedPIFu is able to generate a clean, natural structure of human body. As we have seen earlier, only PIFuHD and our method can reconstruct facial features of the human mesh in great details. However, PIFuHD remains prone to generating noisy, wavy artefacts on the body of reconstructed meshes, and only our method can reconstruct the ears of the human body mesh properly. Moreover, in the fourth row, we observe that only IntegratedPIFu managed to produce a non-broken, naturally-looking wrist. 

We also provide a qualitative evaluation of our method against PIFuHD using real Internet photos in our supplementary materials.

\vspace{-3mm}

\subsubsection{Quantitative Evaluation}
In addition, we show a quantitative evaluation of our method against the state-of-the-arts methods in Tab. \ref{table:sotaQuantitative}. 

In the first five rows of In Tab. \ref{table:sotaQuantitative}, we compare the state-of-the-art methods with ours without the use of predicted depth or human parsing maps. We exclude these maps to allow for a fairer assessment. As seen from the table, with HRI and DOS, we are able to outperform the existing methods in all metrics. 

In addition, in the last three rows of Tab. \ref{table:sotaQuantitative}, we also show the results of incorporating depth (D) or human parsing (P) maps. The results seems to suggest that our approach works best when only human parsing map is given or when neither of the maps are given. However, our ablation studies later will show that IntegratedPIFu with both predicted depth maps and human parsing maps would be a more robust strategy. We show there that using only human parsing maps or neither of the maps would lead to broken limbs or floating artefacts around reconstructed meshes. These phenomena are not obvious if we consider only the quantitative metrics.

\vspace{-4mm}
\setlength{\tabcolsep}{4pt}
\begin{table}
\begin{center}
\caption{Quantitative evaluation of our methods against the state-of-the-arts in the THuman2.0 test set and BUFF dataset. (HRI=High-resolution Integrator, DOS=Depth-Oriented Sampling, D=depth, P=human parsing)}
\vspace{-4mm}
\label{table:sotaQuantitative}
\resizebox{\columnwidth}{!}{
\begin{tabular}{llll|lll}
\hline 
 &\multicolumn{3}{c|}{THuman2.0} & \multicolumn{3}{c}{BUFF} \\
Methods & CD (10\textsuperscript{-4}) & P2S (10\textsuperscript{-4}) & Normal & CD (10\textsuperscript{3}) & P2S (10\textsuperscript{3}) & Normal \\
\hline
Geo-PIFu & 5.816 & 9.657 & 2.452 & 6.250 & 10.757 & 2.912 \\ 
PIFu & 3.135 & 3.072 & 1.731 & 2.639 & 3.367 & 1.947 \\
PIFuHD & 2.800 & 2.540 & 1.698 & 2.031 & 2.029 & 2.010 \\ 
Ours-  HRI & 2.841 & 2.177 & {\bf 1.622} & 2.083 & 1.931 & {\bf 1.762} \\ 
Ours-  HRI+DOS & {\bf 2.711} & {\bf 2.139} & 1.643 & {\bf 2.029} & {\bf 1.925} & 1.800 \\ 
\hline
Ours-  HRI+DOS (D+P)  & 3.040 & 2.259 & 1.641 & 2.287 & 2.224 & 1.842 \\
Ours-  HRI+DOS (D) & 3.003 & 2.387 & {\bf 1.633} & 2.182 & 2.094 & 1.858 \\
Ours-  HRI+DOS (P) & {\bf 2.734} & {\bf 2.153} & 1.780 & {\bf 2.030} & {\bf 1.897} & {\bf 1.789} \\
\hline
\end{tabular}
}
\end{center}
\vspace{-14mm}
\end{table}
\setlength{\tabcolsep}{1.4pt}

\subsection{Ablation Studies}

\subsubsection{Comparison of different Backbones in PIFuHD}
\begin{figure}[t]
\centering
\includegraphics[width=0.9\textwidth]{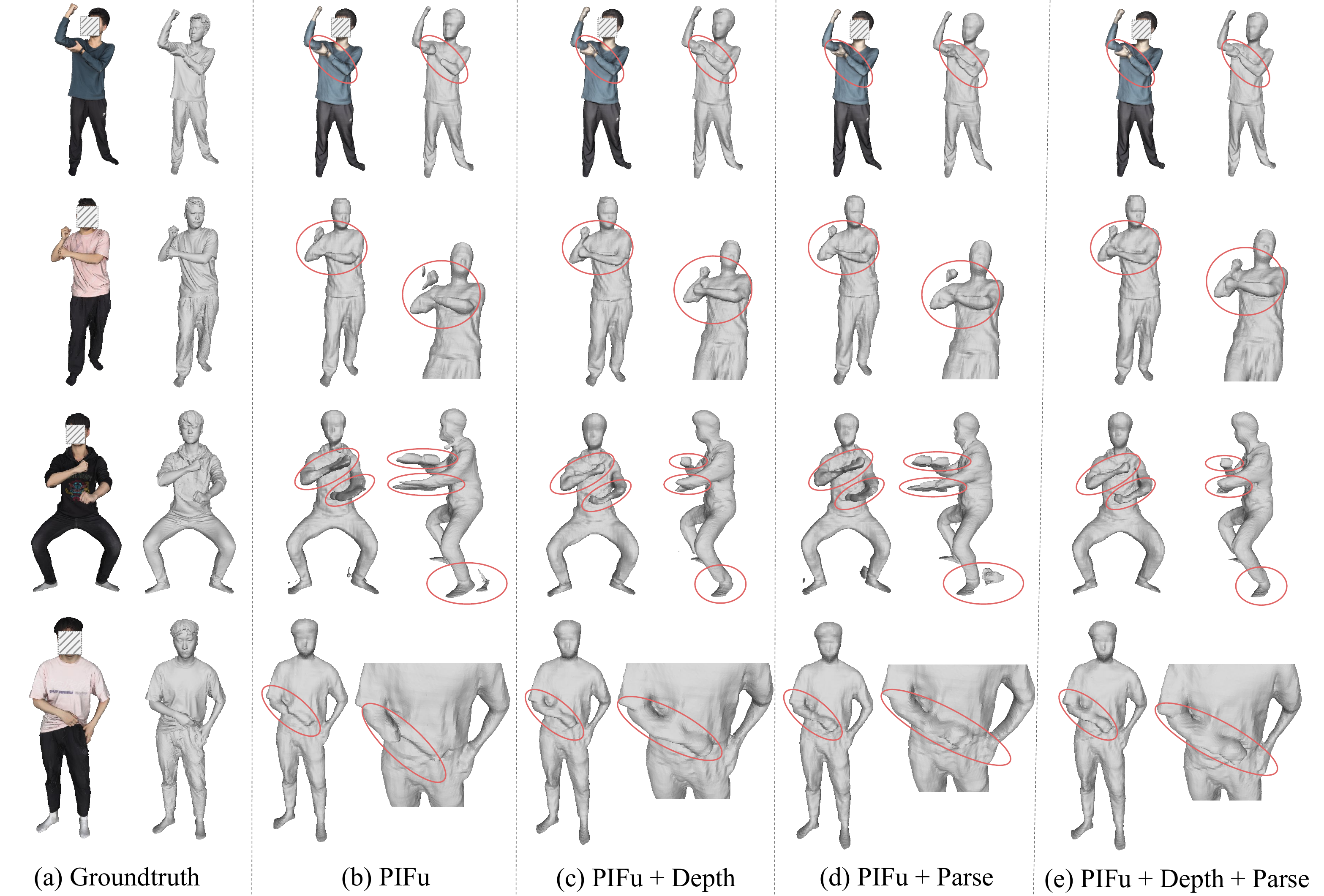}
\vspace{-4mm}
\caption{Comparison with different backbones. (b) is a vanilla PIFu, (c) is given predicted depth map as additional input, (d) is given predicted human parsing, (e) is given both maps. To aid visualization, some reconstructed meshes are colored by projecting the RGB values from the input image onto the generated meshes}
\label{fig:backboneComparison}
\vspace{-5mm}
\end{figure}

Both PIFuHD and IntegratedPIFu use a low-resolution PIFu (also called the ``backbone") to provide a coarse structure of a human mesh. IntegratedPIFu incorporates predicted depth and human parse map into its backbone. In order to evaluate the impact of these maps, we try out different possible backbones and show the results in Fig. \ref{fig:backboneComparison}. 

In the first row of the figure, we observe that (b) a PIFu with neither predicted depth nor human parsing maps (i.e. vanilla PIFu) produced a left arm with a very noticeable breakage. But when either (c) predicted depth, (d) human parsing, or (e) both maps are added to the PIFu module, the left arm is reconstructed properly without any breakage. Reconstructed meshes of the vanilla PIFu have similar breakage in the second and fourth rows in the figure. In the third row, vanilla PIFu produces floating artefacts near the foot of the reconstructed mesh. Furthermore, the reconstructed arms are unnaturally elongated. All of these problems are not replicated by both (c) and (e). 

We find that while (d) a PIFu with human parse map only may mitigate the problems, it does not always solve all of them. Qualitatively, we find that (e) produce the most natural-looking human meshes (explained in supplementary materials). Also, we observe that errors in the backbone would usually be propagated to the PIFuHD (or IntegratedPIFu). To show this, we deliberately pick the same human model for the third row of Fig. \ref{fig:backboneComparison} and the first row of Fig. \ref{fig:secondPageSota}. When the backbone of PIFuHD erroneously produced elongated human arms, the PIFuHD would also replicate this error. More results in supp. materials.

\subsubsection{Evaluation of Depth-Oriented Sampling}
\begin{figure}[h]
\centering
\includegraphics[width=0.9\textwidth]{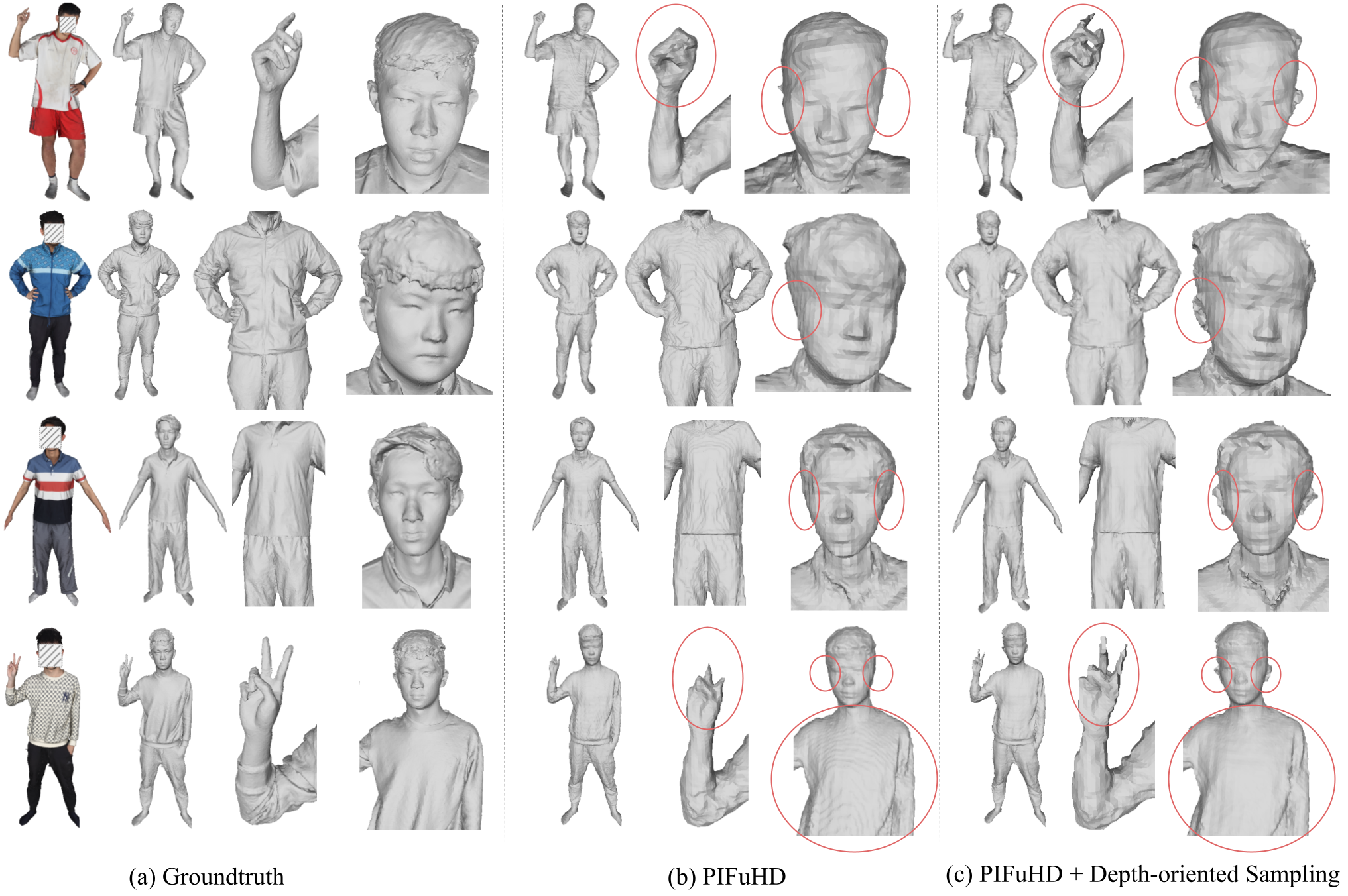}
\vspace{-4mm}
\caption{Evaluation of the effect of depth-oriented sampling. Depth-oriented sampling captures small important features (e.g. fingers, ears) without producing wavy artefacts}
\label{fig:depthOrientedComparison}
\vspace{-5mm}
\end{figure}

In order to evaluate the impact of using depth-oriented sampling (DOS), we train a PIFuHD using the original spatial sampling scheme and compare that with a PIFuHD trained with DOS. Neither models are given depth or human parsing maps. We evaluate the two PIFuHDs in Fig. \ref{fig:depthOrientedComparison}. From the figure, we see the two advantages DOS has over spatial sampling scheme. The first is capturing small but important human features such as fingers and ears. The second advantage is that, unlike spatial sampling, DOS does not produce wavy artefacts on the reconstructed meshes' body. 

\vspace{-3mm}

\subsubsection{Evaluation of High-Resolution Integrator}
\begin{figure}[h]
\vspace{-4mm}
\centering
\includegraphics[width=0.9\textwidth]{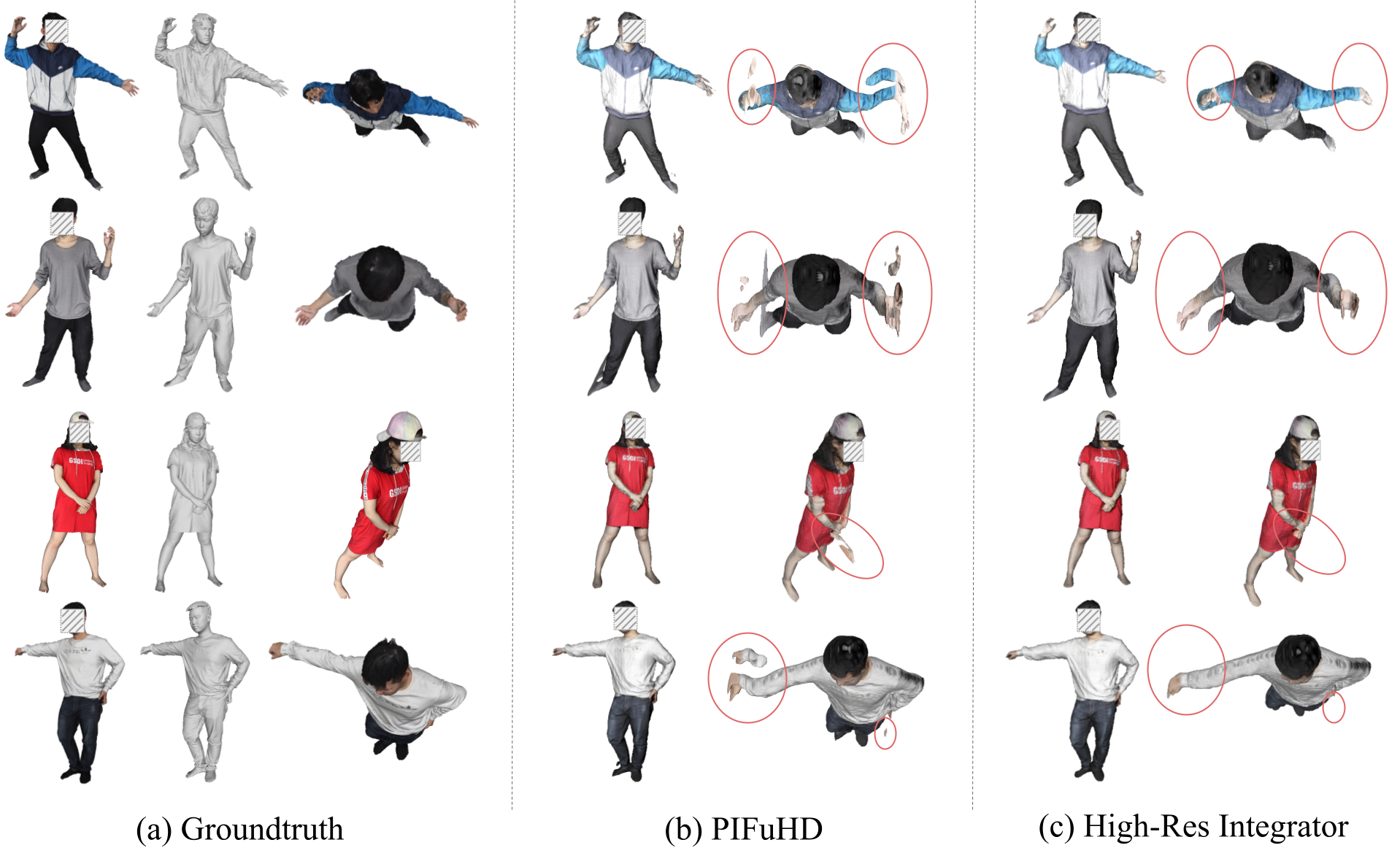}
\vspace{-4mm}
\caption{Comparison of PIFuHD with our HRI. Coloring serves as visual aids only}
\label{fig:integratedPIFuComparison}
\vspace{-14mm}
\end{figure}

In order to evaluate our High-Resolution Integrator (HRI), we train a vanilla PIFuHD and a PIFuHD that has its high-PIFu replaced with our HRI. Both models use the original spatial sampling scheme and are not given depth or human parsing maps. We present the results on Fig. \ref{fig:integratedPIFuComparison}. We find that a PIFuHD modified with our HRI is better at capturing the structure of human meshes compared to the vanilla PIFuHD, which often produces floating artefacts.

\vspace{-3mm}
\section{Conclusion}
\vspace{-3mm}
We have presented three novel techniques to improve any existing pixel-aligned implicit models. Firstly, we showed how relative depth and human parsing information can be predicted and incorporated into a pixel-aligned implicit model. Next, we described depth-oriented sampling, a new training scheme that gives more precise supervision signals than the original spatial sampling scheme. Finally, we introduce High-resolution Integrator, a new architecture that can work in closer tandem with the backbone to reconstruct structurally accurate human meshes. Together, these techniques form our IntegratedPIFu framework.

\vspace{-3mm}
\subsubsection{Acknowledgements} This study is supported under the RIE2020 Industry Alignment Fund – Industry Collaboration Projects (IAF-ICP) Funding Initiative, as well as cash and in-kind contribution from the industry partner(s). G. Lin's participation is supported by the Ministry of Education, Singapore, under its Academic Research Fund Tier 2 (MOE-T2EP20220-0007).

\vspace{-5mm}

\clearpage
%
%
\bibliographystyle{splncs04}
\bibliography{eccv2022submission}
\end{document}